\documentclass[letterpaper, 10 pt, conference]{ieeeconf}  

\IEEEoverridecommandlockouts                              

\usepackage[hidelinks]{hyperref}
\usepackage[sorting=none]{biblatex}
\hypersetup{
  colorlinks   = true, 
  urlcolor     = blue, 
  linkcolor    = blue, 
  citecolor   = blue 
}

\usepackage{graphicx, color}
\usepackage{multirow}
\usepackage{algorithm,algpseudocode}
\usepackage{amsmath}
\usepackage{subcaption}

\usepackage[table,xcdraw]{xcolor}

\title{\LARGE \bf
Study on Aspect Ratio Variability toward Robustness of Vision Transformer-based Vehicle Re-identification  
}
\author{Mei Qiu, Lauren Christopher, and Lingxi Li
\thanks{Mei Qiu, Lauren Christopher, and Lingxi Li  are with the Department of Electrical and Computer Engineering, Purdue University in Indianapolis, 723 West Michigan Street, SL-160, Indianapolis, Indiana 46202, USA. Emails:
        {\tt\small \
        meiqiu@iu.edu,\{lauchris,ll7\}@iupui.edu.
        }}%
}

\begin{document}

\definecolor{baselinecolor}{gray}{.9}
\newcommand{\baseline}[1]{\cellcolor{baselinecolor}{#1}}

\def\sign{\texttt{sign}}
\def\ORAT{\texttt{ORAT}}

\maketitle
\pagestyle{plain}

\begin{abstract}
Vision Transformers (ViTs) have excelled in vehicle re-identification (ReID) tasks. However, non-square aspect ratios of image or video input might significantly affect the re-identification performance. To address this issue, we propose a novel ViT-based ReID framework in this paper, which fuses models trained on a variety of aspect ratios. Our main contributions are threefold: (i) We analyze aspect ratio performance on VeRi-776 and VehicleID datasets, guiding input settings based on aspect ratios of original images. (ii) We introduce patch-wise mixup intra-image during ViT patchification (guided by spatial attention scores) and implement uneven stride for better object aspect ratio matching. (iii) We propose a dynamic feature fusing ReID network, enhancing model robustness. Our ReID method achieves a significantly improved mean Average Precision (mAP) of \textbf{91.0\%} compared to the the closest state-of-the-art (CAL) result of 80.9\% on VehicleID dataset.


\end{abstract}

\section{INTRODUCTION}
\begin{figure}[h!]
    \centering
\includegraphics[width=0.45\textwidth]{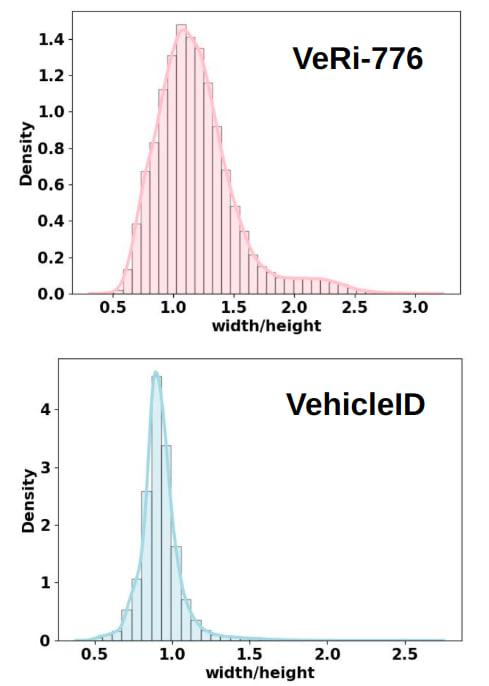} 
    \caption{\textit{
Aspect ratio distribution of images in training datasets from ReID benchmark datasets, VeRi-776 and VehicleID, varies significantly. These datasets show that a substantial portion of the images are non-square.
    }}
    \label{fig:ar_veri_veh}
\end{figure}
A fundamental task in intelligent transportation systems is vehicle re-identification (Re-ID): identifying vehicles across multiple non-overlapping cameras \cite{wang2019survey}. Despite its significance, vehicle Re-ID encounters challenges due to variations in vehicle appearance across different viewpoints, poses, illumination, and backgrounds. Deep learning models face the challenge of extracting discriminative features resistant to viewpoint variations \cite{zheng2020vehiclenet,chen2023global}. Both global and local features are essential for generating robust representations for vehicle pairs \cite{peng2019learning, zhang2020part, gu2021efficient}.

Numerous benchmark datasets contain images from real-world surveillance scenes, including VeRi-776 \cite{zheng2020vehiclenet}, PKU-VD \cite{yan2017exploiting}, VehicleID \cite{liu2016deep}, Vehicle-1M \cite{guo2018learning}, and VERI-Wild \cite{lou2019veri}, crucial for vehicle Re-ID AI development \cite{zakria2021trends}. State-of-the-art models leverage the self-attention mechanism. Vision transformers (ViTs) have demonstrated superior ability in capturing discriminative details compared to previous CNN-based methods \cite{he2021transreid}. Additionally, images in different datasets exhibit various aspect ratios, as illustrated in Fig. \ref{fig:ar_veri_veh}. VeRi-776 and VehicleID display different size and shape distributions, posing a significant challenge in model training.

However, unlike CNN-based models, which can handle varying aspect ratios to some extent due to their translation invariance and local receptive fields, vision transformers consider the entire image as a sequence of patches. The fixed patch size and sequential nature of transformers necessitate careful consideration of how input images are resized and cropped \cite{ke2021musiq, liu2022aspect, mao2022towards, dehghani2023patch}. Early implementations of vision transformers adopted resizing strategies from CNNs, often distorting the original aspect ratio and potentially compromising performance, particularly in tasks reliant on object shape and scale.

To address this, subsequent studies explored padding strategies, adaptive post-patch extraction, trainable resizing networks, aspect-ratio aware attention mechanisms, and multi-scale/multi-aspect training \cite{xia2022vision, lv2022scvit, hwang2022vision, zhu2022aret, ke2021musiq, li2022multi}. However, these approaches entail computational burdens, data requirements, and optimization challenges.

In summary, gaps persist in applying Vision Transformers (ViTs) to vehicle Re-ID. Optimization of scaling and resizing strategies, understanding aspect ratio effects, and exploring data augmentations such as mixup \cite{zhang2018mixup} at the patch level for ViTs in multi-aspect ratio scenarios are needed. Intra-image mixup can potentially enhance the model's ability to learn detailed features, particularly when the whole image is distorted by resizing with unsatisfactory aspect ratios. TransReID \cite{he2021transreid} introduces a jigsaw patch module (JPM) to enhance feature robustness and discrimination ability, representing a step towards addressing these challenges. However, this intra-mixup occurs at the feature level; its potential at the pixel level needs to be comprehensively explored.

\begin{figure}
    \centering
\includegraphics[width=0.48\textwidth]{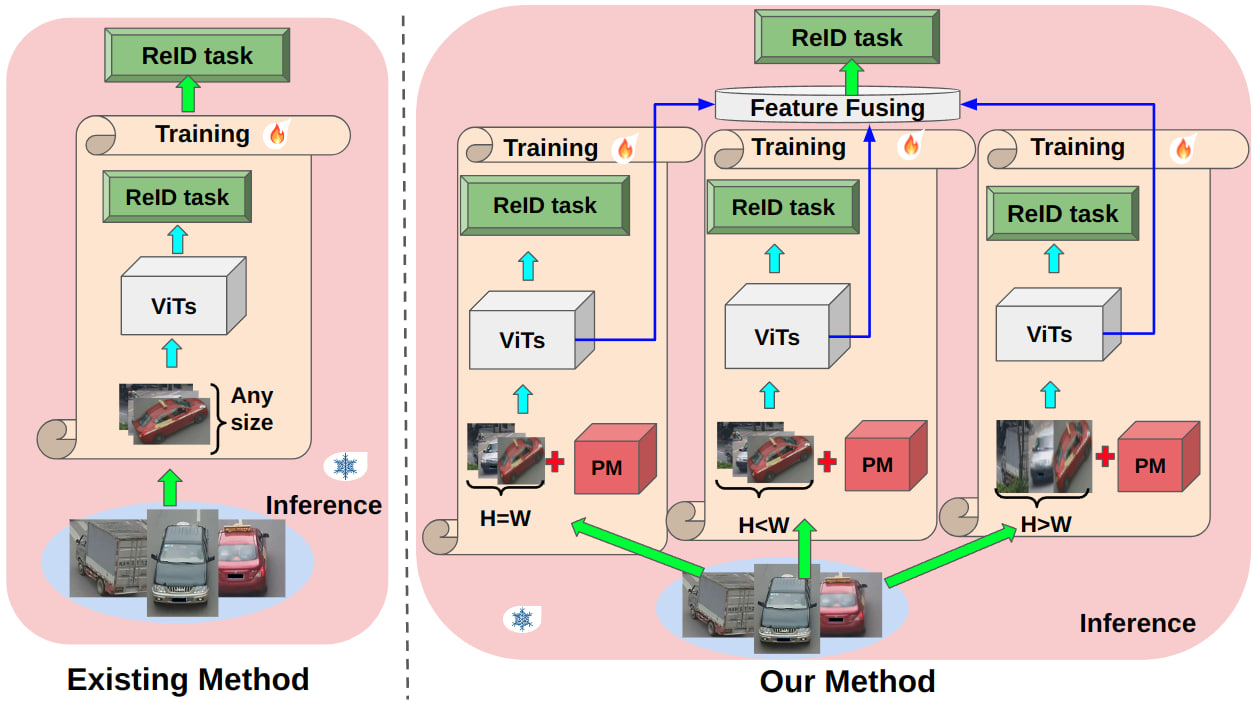} 
    \vspace{-5mm}
    \caption{\textit{ (Left) \textbf{Existing Method}: Image size is typically fixed and set to a single size with a square shape. (Right) \textbf{Our Method}: Combined Vision Transformer (ViT)-based ReID model that dynamically fuses features extracted from multiple models. Each model is trained on a fixed size and aspect ratio.
    }}
    \label{fig:contrib}
    \vspace{2mm}
\end{figure}
\begin{figure*}
    \centering
\includegraphics[width=0.9\textwidth]{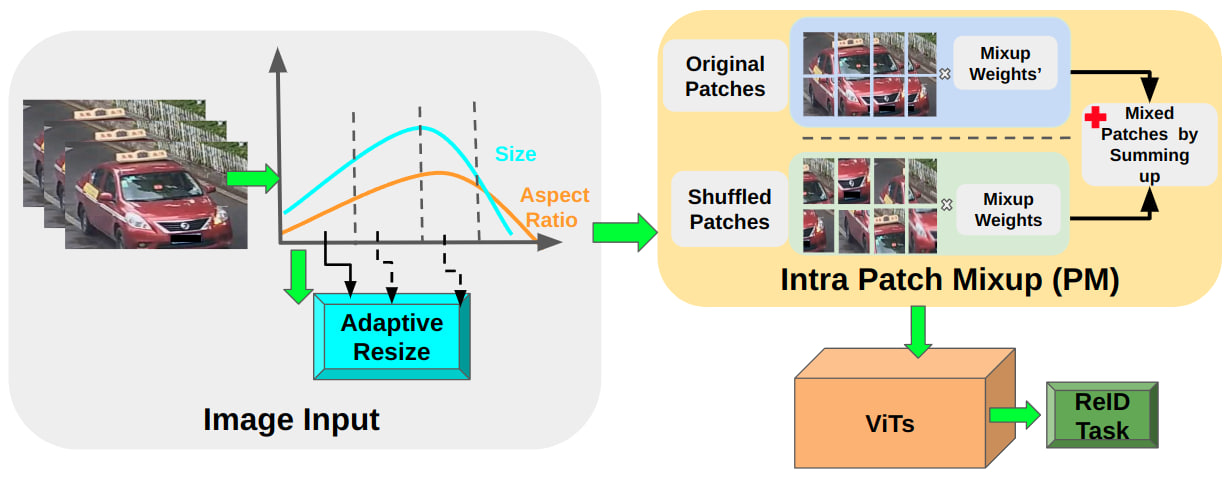} 
    \vspace{1mm}
    \caption{\textit{The structure of each individual model is designed to adapt to the dataset's size and aspect ratio distribution. During the patchification process, the stride size in the horizontal and vertical directions is dynamically determined based on the input object's aspect ratio. Subsequently, a Patch Mixing (PM) module shuffles and mixes patches from the same image using an attention-guided strategy. Any Vision Transformer (ViT)-based architecture can be chosen as the backbone. In this study, we select ViT/B-16. The features extracted from ViTs are used for the vehicle ReID downstream task.}}
    \label{fig:contrib2}
    \vspace{2mm}
\end{figure*}
In this work, we are the first to propose that aspect ratio is a key factor affecting vehicle Re-ID performance and the robustness of feature learning in Vision Transformers (ViTs). We conduct a series of novel, scientific, and comprehensive experiments to explore the effects of various aspect ratios on ViT-based ReID.

To enhance the model's generality to various aspect ratios in the input, we dynamically fuse features extracted from several models trained on images with different aspect ratios, as shown in Fig. \ref{fig:contrib}. Additionally, when training a single model, we propose a novel intra-image patch mixup (PM) data augmentation method to improve the model's learning ability on details and mitigate overfitting during training. Furthermore, to mitigate the distortion caused by unsatisfactory resizing, we employ an uneven stride strategy in the patchify step.

The key contributions of this work are:

\begin{itemize}
    \item Proposed using aspect ratio as a critical factor impacting vehicle Re-ID performance and ViT's feature learning robustness.
    \item Conducted novel and comprehensive experiments to explore the effects of various aspect ratios on ViT-based ReID.
    \item Introducted dynamic feature fusion from models trained on different aspect ratios to enhance model generality.
    \item Proposed an intra-image patch mixup (PM) data augmentation method to improve model learning ability and prevent overfitting.
    \item Implementated an uneven stride strategy to reduce distortion in unsatisfactory resizing.
\end{itemize}


\section{Related Works.}
\smallskip
\noindent
\textbf{Vision Transformer.} The Vision Transformer (ViT) is an adaptation of the transformer architecture from natural language processing (NLP) tasks \cite{vaswani2017attention} to computer vision. Dosovitskiy et al. introduced the ViT \cite{dosovitskiy2010image}, the first to demonstrate that a pure transformer applied directly to sequences of image patches can excel in large-scale image classification tasks. Within ViTs, the multi-head self-attention mechanism enables the model to capture diverse dependencies in images, including shapes, textures, and contextual relationships between objects in parallel, facilitating efficient learning of richer data representations. ViTs and their variants, such as DeiT \cite{touvron2021training}, Swin Transformer \cite{liu2021swin}, PVT \cite{wang2021pyramid}, and CPVT \cite{chu2021conditional}, prove beneficial across tasks including image classification, object detection, and Re-ID \cite{khan2022transformers}.

\smallskip
\noindent
\textbf{Vehicle Re-Identification.} As discussed, vehicle Re-ID constitutes the primary focus of this study. Numerous works leveraging deep learning have achieved notable performance across various public vehicle Re-ID benchmarks. These works often utilize either CNN backbones \cite{he2016deep, bashir2019vr, roman2021improving} or ViT backbones \cite{he2021transreid, lian2022transformer, luo2021empirical, wei2022transformer}. Common loss functions employed in deep vehicle Re-ID network training include cross-entropy loss (ID loss) \cite{zheng2017discriminatively}, triple loss \cite{liu2017end}, and contrastive loss \cite{hadsell2006dimensionality}.






\begin{figure}
    \centering
\includegraphics[width=0.48\textwidth]{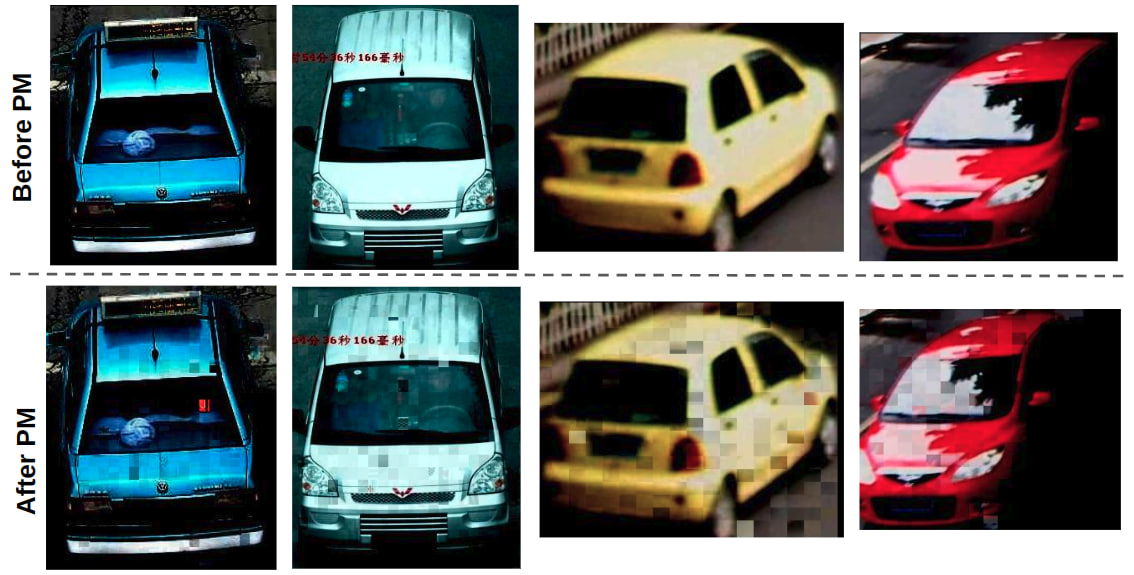} 
    \vspace{-5mm}
    \caption{\textit{Displayed are examples from the VehicleID (first two columns) and VeRi-776 (last two columns) test datasets, illustrating the effects of the intra-image patch mixup (PM) data augmentation method. This technique blends various parts of an image based on attention-driven distances, increasing image complexity to enhance model robustness and reduce overfitting. The top row presents images without the PM module, while the bottom row features images processed with the PM module.
    }}
    \label{fig:aug}
    \vspace{2mm}
\end{figure}

\section{Method}
\smallskip
\noindent
\textbf{Model Structure.} We train one model for each major aspect ratio. The determination of how many models are needed for training with various aspect ratios is learned from the data, as illustrated in Fig. \ref{fig:contrib2}. For each model with a fixed aspect ratio, the resized input images undergo augmentation using a Patch Mixup (PM) module. Subsequently, a chosen Vision Transformer (ViT) backbone and its pre-trained weights are used for initialization. Features extracted from the last transformer layer are utilized for the ReID task.

\smallskip
\noindent
\textbf{Image Input with Adaptive Size and Shape.} We employ a statistical method to estimate the dataset's size and aspect ratios. Starting with the mean or median value of the dataset's size is a suitable approach. For aspect ratios, clustering methods such as K-means can be used to generate several clusters representing the aspect ratios present in the original dataset. Subsequently, a combination of size and aspect ratio guides the resizing of images for the training of each specific model. Each model is trained on a fixed size and aspect ratio.

\smallskip
\noindent
\textbf{Patchification with uneven stride.}
To enhance the model's ability to learn spatial relationships from image data, we incorporate uneven strides in different dimensions, based on the aspect ratio. Specifically, in our approach, the stride sizes $s_h$ and $s_w$ can vary, with the stride size being smaller in the shorter dimension compared to the longer dimension. We maintain a fixed patch size $p$ of 16. The total patch number $n$ is calculated as $\left(\frac{H - p}{s_h} + 1\right) \times \left(\frac{W - p}{s_w} + 1\right)$.




   

\smallskip
\noindent
\textbf{Patch Mixup intra image module.} We propose a novel intra-image data augmentation method where each patch of an image has a probability to mix adaptively with another randomly chosen patch from the same image. Mixup weights are determined based on the spatial distance between these two patches in pixels, with closer patches assigned higher mixup weights. Let $A$ denote the original patches with $n$ patches, and their positions are denoted by the coordinates of their top-left corners $(x_i, y_i)$ for the $i$-th patch. After patch partitioning, all patch indexes are fixed, with each patch having a unique index ranging from $0$ to $n-1$. The Euclidean distance between the centers of two patches indexed by $i$ and $j$ can be computed as:
\begin{equation}
d(i, j) = \sqrt{(x_i - x_j)^2 + (y_i - y_j)^2}
\label{d_dis}
\end{equation}

The patch distance matrix \( D \) for all patches can then be represented as:

\begin{equation}
D = \begin{bmatrix}
d(1, 1) & d(1, 2) & \cdots & d(1, n) \\
d(2, 1) & d(2, 2) & \cdots & d(2, n) \\
\vdots & \vdots & \ddots & \vdots \\
d(n, 1) & d(n, 2) & \cdots & d(n, n)
\end{bmatrix}
\label{d_dis_matrix}
\end{equation}

Given a distance matrix \(D\) where each element \(D_{ij}\) represents the distance between patch \(i\) and patch \(j\), the attention scores matrix \(S\) can be computed as:
\begin{equation}
S = \frac{1}{1 + D*p}
\label{attention_score}
\end{equation}

In this formula, the attention score $S_{ij}$ increases as the distance $D_{ij}$ decreases, indicating that closer patches have more influence on each other.
After shuffling patches, we generate a new patch-to-patch mixup weights matrix $S'$ based on the patches' indexes, as shown in Equation \ref{attention_score-adjusted}, where each value of $S'$ is fetched from $S$ using the indexes of the patches:
\begin{equation}
S' = S[A_k, B_k']
\label{attention_score-adjusted}
\end{equation}
Here, $k$ is the patch index of a patch from the original patches $A$, and $k'$ is the patch index of the selected patch from the shuffled patches $B$ for mixup processing. 
Assuming the original patches $A$ are generated using non-overlapping patchification, and its patch distance matrix $D$ is calculated based on Equation \ref{d_dis} and \ref{d_dis_matrix}, with $D = 16 \times D$ as the patch is a fixed-size square. The original spatial attention scores and mixup weights matrix based on the shuffled patches are calculated using Equations \ref{attention_score} and \ref{attention_score-adjusted}.
The final patches of this image $F$ are given by:
$F = S' \times B + (1-B) \times A$. 
Several image samples before and after Patch Mixing (PM) data augmentation are shown in Fig. \ref{fig:aug}, where it can be observed that after PM, detailed information from different parts of the same vehicle is mixed. The ablation study result is shown in Table \ref{tab:basci-res}. The pseudocode of this method is shown in Algorithm \ref{alg:patch_mix}.
\begin{algorithm}
\caption{Patch Mixing guided by Spatial Attention Scores in Vision Transformer}\label{alg:patch_mix}
\begin{algorithmic}[1]
\Require An input image $I$ of dimensions $H \times W \times C$, a patch size $P$, and stride $S$ for overlapping.
\State $B \gets \text{batch size of } I$
\State $N \gets \text{number of patches along height}$
\State $M \gets \text{number of patches along width}$
\State $D \gets \text{patch distance matrix of size } (N \times M) \times (N \times M)$
\State $S \gets \text{compute attention scores from } D$ 
\For {$b = 1$ to $B$}
    \State $P_b \gets \text{extract patches from image } I_b \text{ with overlap}$
    \State $\pi \gets \text{random permutation of } \{1, \ldots, N \times M\}$
    \State $P'_b \gets P_b[\pi] \text{ shuffled patches using permutation } \pi$
    \State $S'_b \gets S[\pi] \text{ adjust attention scores with } \pi$
    \For {$i = 1$ to $N \times M$}
        \State $\lambda \gets S'_{b_{i, \pi(i)}} \text{ attention weight for current patch}$
        \State $P''_{b_i} \gets (1-\lambda) \cdot P_{b_i} + \lambda \cdot P'_{b_{\pi(i)}}$
    \EndFor
    \State $O_b \gets \text{reconstruct output from mixed patches } P''_b$
\EndFor
\State \textbf{return} $O$ \Comment{Return the batch of mixed images}
\end{algorithmic}
\end{algorithm}

\smallskip
\noindent
\textbf{ReID Task.}
We optimize the network by constructing ID loss and triplet loss for global features. The ID loss $L_{ID}$ is the cross-entropy loss without label smoothing. For a triplet set $\{a, p, n\}$, the triplet loss $L_T$ with a soft-margin. 
The triplet loss $L_T$ with soft-margin is defined as:

\[
L_T = \log(1 + \exp(||\mathbf{f}_a - \mathbf{f}_p||^2 - ||\mathbf{f}_a - \mathbf{f}_n||^2))
\]

where:
\begin{itemize}
  \item $\mathbf{f}_a$, $\mathbf{f}_p$, and $\mathbf{f}_n$ are the feature embeddings of the anchor, positive, and negative samples, respectively.
  \item $||\cdot||^2$ denotes the squared Euclidean distance.
\end{itemize}

\smallskip
\noindent
\textbf{Inference Phase.} As shown in Fig. \ref{fig:contrib}, we use a \textbf{Dynamic Feature Fusing} strategy during inference to fuse features from models trained on multiple inputs. The output feature vector can be represented as:
\[
\textbf{F\_out} =  w_1 \cdot \textbf{f}_1 + w_2 \cdot \textbf{f}_2 +  w_3 \cdot \textbf{f}_3 
\]
or as a weighted concatenation:
\[
\textbf{F\_out} = \left[ w_1 \cdot \textbf{f}_1 ; w_2 \cdot \textbf{f}_2; w_3 \cdot \textbf{f}_3 \right]
\]

where \(w_1\), \(w_2\), and \(w_3\) are weights. For the adaptive weight assignment \(w\):

\[
w = 
\begin{cases}
    1.3, & \text{if } |\text{model\_ar} - \text{image\_ar}| \leq 0.3 \\
    1.0, & \text{if } 0.3 < |\text{model\_ar} - \text{image\_ar}| \leq 0.6 \\
    0.9, & \text{otherwise}
\end{cases}
\]

We investigated the ReID performance by summing or concatenating features, and the results are shown in Table \ref{tab:basci-res}.

\section{Experiments}
\subsection{Experiment Settings}
\smallskip
\noindent
\textbf{Dataset.} Two popular vehicle Re-ID benmark datasets: VeRi-776 \cite{zheng2020vehiclenet} and VehicleID \cite{liu2016deep}, are used in this work.



\smallskip
\noindent
\textbf{Evaluation Metrics.} In line with the Re-ID field, we employ mAP (mean Average Precision) and CMC (Cumulative Matching Characteristic) as our performance estimation metrics. Specifically, we consider Rank-1 (R1), Rank-5 (R5), and Rank-10 (R10) accuracy.



\begin{table*}[]
\centering
\resizebox{\textwidth}{!}{%
\begin{tabular}{|l|c|c|c|c|c|c|c|}
\hline
\textbf{Dataset} &
  \textbf{\begin{tabular}[c]{@{}c@{}}Image Size/aspect ratio\\ (w/PM or w/o PM)\end{tabular}} &
  \textbf{Stride Size} &
  \textbf{Fusing Strategy} &
  \textbf{mAP(\%)} &
  \textbf{R1(\%)} &
  \textbf{R5(\%)} &
  \textbf{R10(\%)} \\ \hline
 &
  {\color[HTML]{3166FF} \textbf{{[}224,224{]}/1.0 (w/o)}} &
  {\color[HTML]{3166FF} \textbf{{[}16,16{]}}} &
  {\color[HTML]{3166FF} \textbf{-}} &
  {\color[HTML]{3166FF} \textbf{74.9}} &
  {\color[HTML]{3166FF} \textbf{95.0}} &
  {\color[HTML]{3166FF} \textbf{97.6}} &
  {\color[HTML]{3166FF} \textbf{98.6}} \\ \cline{2-8} 
 &
  {[}224,224{]}/1.0 (w/) &
  {[}16,16{]} &
  - &
  77.8 &
  96.1 &
  98.8 &
  99.1 \\ \cline{2-8} 
 &
  {[}224,212{]}/0.95 (w/o) &
  {[}16,12{]} &
  - &
  75.5 &
  95.2 &
  97.9 &
  98.5 \\ \cline{2-8} 
 &
  {[}224,212{]}/0.95 (w/) &
  {[}16,12{]} &
  - &
  73.2 &
  93.5 &
  97.4 &
  98.4 \\ \cline{2-8} 
 &
  {[}224,298{]}/1.33 (w/o) &
  {[}12,16{]} &
  - &
  78.5 &
  96.5 &
  98.0 &
  99.0 \\ \cline{2-8} 
 &
  {[}224,298{]}/1.33 (w/) &
  {[}12,16{]} &
  - &
  79.4 (4.5$\uparrow$)&
  96.1 &
  98.2 &
  99.0 \\ \cline{2-8} 
 &
  \begin{tabular}[c]{@{}c@{}}Fusing three models \\ (w/o)\end{tabular} &
  - &
  Weighted Sum &
  78.1 &
  96.1 &
  98.1 &
  98.7 \\ \cline{2-8} 
 &
  {\color[HTML]{CB0000} \begin{tabular}[c]{@{}c@{}}Fusing three models\\  (w/)\end{tabular}} &
  {\color[HTML]{CB0000} -} &
  {\color[HTML]{CB0000} Weighted Sum} &
  {\color[HTML]{CB0000} \textbf{81.4}(6.5$\uparrow$)} &
  {\color[HTML]{CB0000} \textbf{97.0(2.0$\uparrow$)}} &
  {\color[HTML]{CB0000} \textbf{98.5}} &
  {\color[HTML]{CB0000} \textbf{99.1}} \\ \cline{2-8} 
 &
  \begin{tabular}[c]{@{}c@{}}Fusing three models\\  (w/o)\end{tabular} &
  - &
  \begin{tabular}[c]{@{}c@{}}Weighted \\ Concatenate\end{tabular} &
  78.0 &
  95.9 &
  98.1 &
  98.6 \\ \cline{2-8} 
\multirow{-10}{*}{VeRi776} &
  \begin{tabular}[c]{@{}c@{}}Fusing three models\\ (w/)\end{tabular} &
  - &
  \begin{tabular}[c]{@{}c@{}}Weighted \\ Concatenate\end{tabular} &
  81.4 &
  97.0 &
  98.5 &
  99.0 \\ \hline
 &
  {\color[HTML]{3166FF} \textbf{{[}384,384{]}/1.0 (w/o)}} &
  {\color[HTML]{3166FF} \textbf{{[}16,16{]}}} &
  {\color[HTML]{3166FF} \textbf{-}} &
  {\color[HTML]{3166FF} \textbf{90.2}} &
  {\color[HTML]{3166FF} \textbf{85.0}} &
  {\color[HTML]{3166FF} \textbf{97.5}} &
  {\color[HTML]{3166FF} \textbf{99.3}} \\ \cline{2-8} 
 &
  {[}384,384{]}/1.0 (w/) &
  {[}16,16{]} &
  - &
  90.8(0.6$\uparrow$) &
  85.7 &
  97.8 &
  99.0 \\ \cline{2-8} 
 &
  {[}384,308{]}/0.80 (w/o) &
  {[}16,12{]} &
  - &
  89.9 &
  84.1 &
  97.6 &
  99.2 \\ \cline{2-8} 
 &
  {[}384,308{]}/0.80 (w/) &
  {[}16,12{]} &
  - &
  89.6 &
  83.8 &
  97.1 &
  98.8 \\ \cline{2-8} 
 &
  {[}384,396{]}/1.03 (w/o) &
  {[}12,16{]} &
  - &
  87.8 &
  81.7 &
  96.1 &
  98.6 \\ \cline{2-8} 
 &
  {[}384,396{]}/1.03 (w/) &
  {[}12,16{]} &
  - &
  90.2 &
  85.2 &
  97.1 &
  98.8 \\ \cline{2-8} 
 &
  \begin{tabular}[c]{@{}c@{}}Fusing three models\\ (w/o)\end{tabular} &
  - &
  Weighted Sum &
  90.1 &
  84.8 &
  97.3 &
  98.9 \\ \cline{2-8} 
 &
  {\color[HTML]{CB0000} \begin{tabular}[c]{@{}c@{}}Fusing three models\\ (w/)\end{tabular}} &
  {\color[HTML]{CB0000} -} &
  {\color[HTML]{CB0000} Weighted Sum} &
  {\color[HTML]{CB0000} \textbf{91.0(0.8$\uparrow$)}} &
  {\color[HTML]{CB0000} \textbf{86.3(1.3$\uparrow$)}} &
  {\color[HTML]{CB0000} \textbf{97.4}} &
  {\color[HTML]{CB0000} \textbf{99.0}} \\ \cline{2-8} 
 &
  \begin{tabular}[c]{@{}c@{}}Fusing three models\\ (w/o)\end{tabular} &
  - &
  \begin{tabular}[c]{@{}c@{}}Weighted \\ Concatenate\end{tabular} &
  90.5 &
  85.4 &
  97.4 &
  99.0 \\ \cline{2-8} 
\multirow{-10}{*}{VehicleID} &
  \begin{tabular}[c]{@{}c@{}}Fusing three models\\ (w/)\end{tabular} &
  - &
  \begin{tabular}[c]{@{}c@{}}Weighted \\ Concatenate\end{tabular} &
  90.7 &
  85.3 &
  97.7 &
  99.1 \\ \hline
\end{tabular}%
}
\caption{The basic results of our method. \textbf{The blue color marks the baseline result we used for comparison and the red color marks the best performance}.}
\label{tab:basci-res}
\end{table*}

\smallskip
\noindent
\textbf{Implementation Details.} After analyzing the data size and aspect ratio distributions from the VeRi-776 and VehicleID training datasets, we standardized the resized image heights to 224 pixels for VeRi-776 and 384 pixels for VehicleID. We used a uniform patch size of $16 \times 16$ for both training and testing phases, with stride sizes set to 16 for the longer dimension and 12 for the shorter. The aspect ratios were $[1.0, 0.95, 1.33]$ for VeRi-776 and $[1.0, 0.80, 1.03]$ for VehicleID. In the testing stage, we used the entire VeRi-776 test dataset and the largest VehicleID subset, containing 800 vehicles. Image preprocessing included 50\% random horizontal flipping, padding, cropping, and erasing.

All experiments were conducted on four NVIDIA RTX A6000 GPUs using PyTorch with FP16 training. We used an SGD optimizer with a momentum of 0.1 and a weight decay of 1e-4, maintaining equal weights of 1.0 for ID and triplet losses. The batch size was set at 128, with four images per ID, across 120 training epochs. Initial learning rates were set at 0.035 for VeRi-776 and 0.045 for VehicleID, both decreasing linearly.

\subsection{Results}


\smallskip
\noindent
\textbf{Major Results of Our Method.} Our experiments (Table \ref{tab:basci-res}) compare baseline performance with different input shapes and the absence of patch mixup (PM) data augmentation. On VeRi-776, non-square input of $224 \times298 $ without PM outperforms square input of $224 \times 224$ by 4.5\% in mAP. With sum-based weighted feature fusion from three models, the mAP improves by over 6.5\%. Similarly, on VehicleID, non-square input of $384\times396$ without PM achieves 0.6\% higher mAP than square input of $384\times384$, while feature fusion raises mAP by over 0.8\%.

\smallskip
\noindent
\textbf{Comparsion with the State-of-the-Art Methods.}
In Tables \ref{tab:com-veri} and \ref{tab:com-veh}, we compare our best model's ReID performance with three state-of-the-art methods (RPTM \cite{ghosh2023relation}, CAL \cite{rao2021counterfactual}, and TransReID \cite{he2021transreid}) on VeRi-776 and VehicleID large test datasets. Our model, using ViT-B/16 backbone, outperforms pure ViT-B/16 by 23\% mAP and 0.5\% R1 on VeRi-776. On VehicleID, our approach achieves 91.0\% mAP and 97.4\% R5, surpassing RPTM by 10.5\% mAP and 1.1\% R5. These results demonstrate the significant potential for improving ReID performance by solely modifying ViTs' input without altering their architectures.

\begin{figure*}[htbp]
  \centering
  \begin{minipage}[b]{0.49\linewidth}
    \centering
    \includegraphics[width=\linewidth]{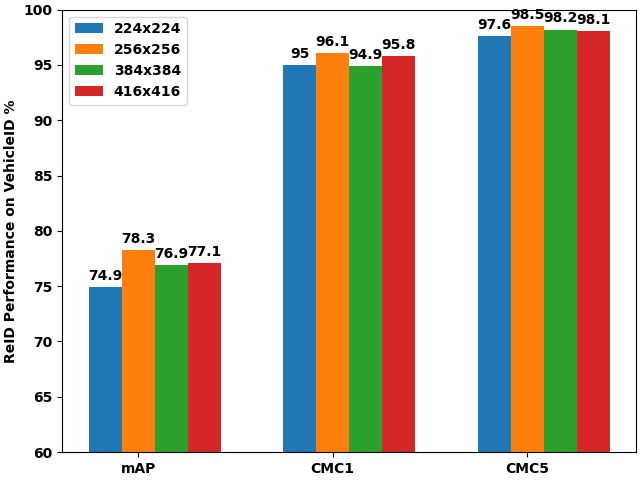}
    \caption{ReID Performance of Square Input on VeRi-776.}
    \label{fig:veri-square}
  \end{minipage}
  \hfill
  \begin{minipage}[b]{0.49\linewidth}
    \centering
    \includegraphics[width=\linewidth]{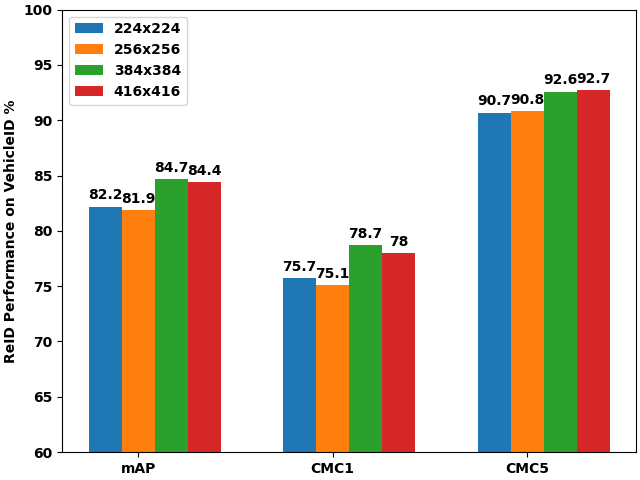}
    \caption{ReID Performance of Square Input on VehicleID.}
    \label{fig:veh-square}
  \end{minipage}
\end{figure*}
\begin{table}[]
\centering
\resizebox{\columnwidth}{!}{%
\begin{tabular}{|c|c|c|c|c|c|}
\hline
\textbf{Backbone} &
  \textbf{\begin{tabular}[c]{@{}c@{}}Input \\ Size\end{tabular}} &
  \textbf{Method} &
  \textbf{mAP} &
  \textbf{R1} &
  \textbf{R5} \\ \hline
ResNet-101 & 240x240 & RPTM\cite{ghosh2023relation}      & 88.0 & 97.3 & 98.4 \\ \hline
ResNet-50  & 384x192 & CAL\cite{rao2021counterfactual}       & 74.3 & 95.4 & 97.9 \\ \hline
ViT/B-16   & 384x128 & TransReID\cite{he2021transreid} & 82.0 & 97.1 & -    \\ \hline
ViT/B-16 &
  256x128 &
  \begin{tabular}[c]{@{}c@{}}ViT/B-16 \\ Baseline\cite{he2021transreid}\end{tabular} &
  78.2 &
  96.5 &
  - \\ \hline
ViT/B-16 &
  Fused &
  \textbf{Ours} &
  {81.5} &
  97.0 &
  {\color[HTML]{CB0000} \textbf{98.5}} \\ \hline
\end{tabular}%
}
\caption{Comparison with state-of-the-art methods of Re-ID on VeRi-776}
\label{tab:com-veri}
\end{table}
\begin{table}[]
\centering
\resizebox{\columnwidth}{!}{%
\begin{tabular}{|c|c|c|c|c|c|}
\hline
\textbf{Backbone} &
  \textbf{\begin{tabular}[c]{@{}c@{}}Input\\  Size\end{tabular}} &
  \textbf{Method} &
  \textbf{\begin{tabular}[c]{@{}c@{}}mAP\\ (\%)\end{tabular}} &
  \textbf{R1(\%)} &
  \textbf{R5(\%)} \\ \hline
ResNet-101 & 240x240 & RPTM\cite{ghosh2023relation}                                                         & 80.5                                 & 92.9 & 96.3 \\ \hline
ResNet-50  & 384x192 & \cite{rao2021counterfactual}                                                          & 80.9                                 & 75.1 & 88.5 \\ \hline
ViT/B-16   & 256x128 & TransReID\cite{he2021transreid}                                                   & -                                    & 85.2 & 97.5 \\ \hline
ViT/B-16   & 256x128 & \begin{tabular}[c]{@{}c@{}}ViT/B-16 \\ Baseline\cite{he2021transreid}\end{tabular} & -                                    & 83.5 & 96.7 \\ \hline
ViT/B-16   & Fused   & {\color[HTML]{FE0000} \textbf{Ours}}                         & {\color[HTML]{CB0000} \textbf{91.0}} & 86.3 & 97.4 \\ \hline
\end{tabular}%
}
\caption{Comparison with state-of-the-art methods of Re-ID on VehicleID large dataset}
\label{tab:com-veh}
\end{table}

\smallskip
\noindent
\textbf{Ablation Study: Square input with different sizes.}
We assess the impact of input size on ReID performance using square shapes of $224 \times 224$, $256 \times 256$, $384 \times 384$, and $416 \times 416$ for both VeRI-776 and VehicleID datasets. All the trainings are conducted without using the PM module. The results are shown in Fig. \ref{fig:veri-square} and \ref{fig:veh-square}. On both datasets, the ReID performance does not linearly increase with the input size. The highest ReID accuracy is observed when the input size is 256 pixels for VeRi-776 and 384 pixels for VehicleID.
\begin{figure}[h!]
    \centering
\includegraphics[width=0.48\textwidth]{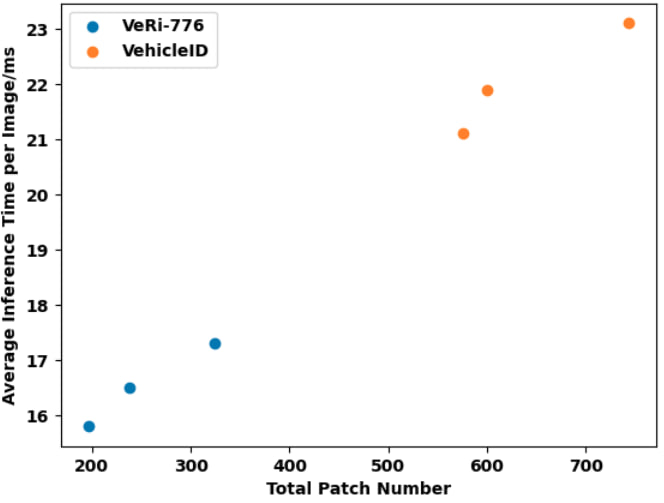} 
    \vspace{-5mm}
    \caption{\textit{Average inference Time of  single Image from Test Dataset.
    }}
    \label{fig:time_cost}
    \vspace{2mm}
\end{figure}

\begin{figure}[h!]
    \centering
\includegraphics[width=0.48\textwidth]{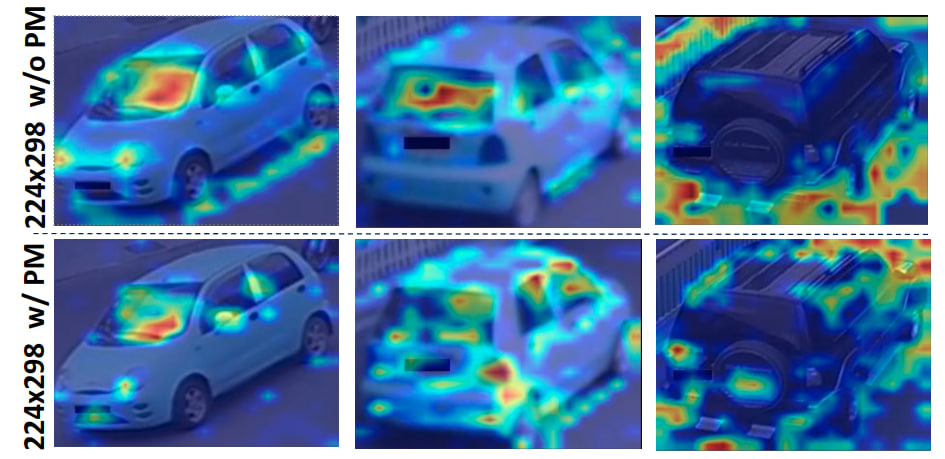} 
    \vspace{-5mm}
    \caption{\textit{Attention map on VeRi-776 without and with PM module. The first row shows the results without using PM module, and the second row shows the results using PM module.
    }}
    \label{fig:atten-veri}
    \vspace{2mm}
\end{figure}

\begin{figure}[h!]
    \centering
\includegraphics[width=0.48\textwidth]{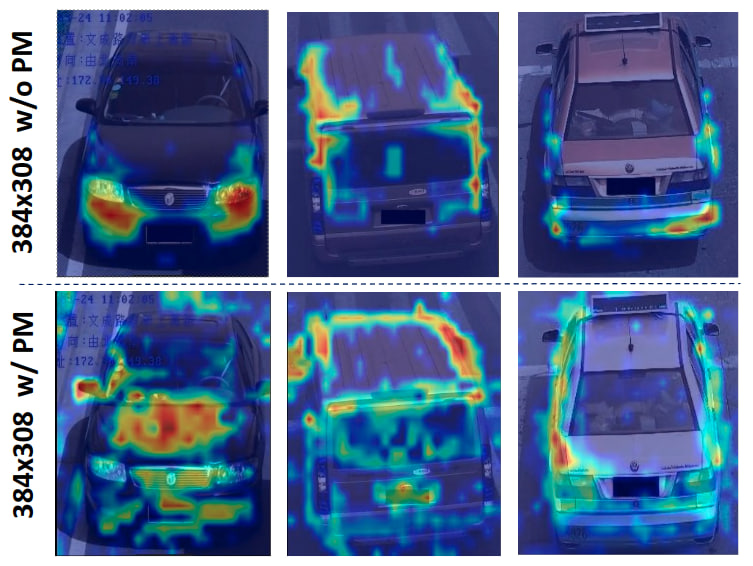} 
    \vspace{-5mm}
    \caption{\textit{Attention map on VehicleID without and  with PM module. The first row shows the results without using PM module, and the second row shows the results using PM module.
    }}
    \label{fig:atten-veh}
    \vspace{2mm}
\end{figure}

\subsection{Results Analysis}

\smallskip
\noindent
\textbf{Patch Number affects Average Inference Time.} Fig. \ref{fig:time_cost} illustrates the correlation between patch count and inference time per object. Increased patches escalate time costs, indicating optimization opportunities. Despite longer inference times from model fusion, mitigation is possible through mask or attention strategies to trim patch count during training, thus curtailing parameter size.

\smallskip
\noindent
\textbf{How does patch mixup (PM) work?} Grad-CAM visualizes attention (Fig. \ref{fig:atten-veri} and \ref{fig:atten-veh}). Intra-image Patch Mixup improves object focus, especially in VeRi-776, enhancing robustness to view and aspect ratios, notably in VehicleID. Utilizing self-attention, our method enhances spatial relations, improving global feature learning in ViTs.


\section{Conclusion }
We explore aspect ratio's impact on ViT-based vehicle ReID. Fusion of varied aspect ratio models boosts robustness and Re-ID performance. Our intra-image Patch Mixup augmentation enhances generalization. Outperforming baselines on VeRi-776 and VehicleID, though fusion may raise inference time, network pruning can be used to mitigate this. Future work aims at efficient ReID models for diverse aspect ratios.

\newpage
\printbibliography

\end{document}